\documentclass[11pt]{article}
\usepackage{geometry}
\usepackage{coling2020}
\usepackage{times}
\usepackage{url}
\usepackage{latexsym}
\usepackage{times}
\usepackage{latexsym}
\usepackage{tabularx}
\usepackage{soul}
\usepackage{times}
\usepackage{latexsym}
\usepackage{multirow}
\usepackage{multicol}
\usepackage{amsmath}
\usepackage{graphicx}
\usepackage{amsfonts}
\usepackage{blkarray}
\usepackage[]{algorithm2e}
\usepackage{array}
\usepackage{booktabs}
\usepackage{footnote}

\usepackage{hyperref}
\usepackage{xstring}

\usepackage{url}
\colingfinalcopy 

\title{Unsupervised Embedding-based Detection of Lexical Semantic Changes}

\author{Ehsaneddin Asgari$^\dag$, Christoph Ringlstetter$^\dag$, Hinrich Sch\"{u}tze$^\diamond$\\
$^\dag$ NLP Expert Center, Data:Lab, Volkswagen AG, Munich, Germany\\
$^\diamond$ Center for Information and Language Processing, LMU Munich, Germany \\
{\tt {\footnotesize \{ehsaneddin.asgari, christoph.ringlstetter\}@volkswagen.de}} \\
{\tt {\footnotesize inquiries@cislmu.org}} \\}

\date{}

\begin{document}
\maketitle
\begin{abstract}
This paper describes EmbLexChange, a system introduced by the ``Life-Language'' team for SemEval-2020 Task 1, on unsupervised detection of lexical-semantic changes. EmbLexChange is defined as the divergence between the embedding based profiles of word $w$ (calculated with respect to a set of reference words) in the source and the target domains (source and target domains can be simply two time frames $t_1$ and $t_2$). The underlying assumption is that the lexical-semantic change of word $w$ would affect its co-occurring words and subsequently alters the neighborhoods in the embedding spaces. We show that using a resampling framework for the selection of reference words, we can reliably detect lexical-semantic changes in English, German, Swedish, and Latin. EmbLexChange achieved second place in the binary detection of semantic changes in the SemEval-2020.\\
\end{abstract}

\section{Introduction}
SemEval 2020 Task 1 is defined on the unsupervised detection of word sense changes over time in German, English, Swedish, and Latin. In particular, this challenge focused on detection and quantification of the sense changes of word $w$ in the transition from the time period $t_1$ to the time period $t_2$ in the above mentioned four languages, where the input for each language are the text corpora dating from $t_1$ and $t_2$. This challenge involved two sub-tasks:\\

\textbf{i. Classification}: The goal of the classification task is the binary detection of lexical semantic change (from $t_1$ to $t_2$) for the given word $w$.\\

\textbf{ii. Ranking}: This sub-task involves the ranking of lexical-semantic change for a given list of words ($w_1, w_2, \dots, w_M$) by assigning scores quantifying relative changes of the word senses. \\

To measure the two sub-tasks, the participating
systems are  evaluated against a ground truth corpus annotated by native speakers 
or scholars of the respective languages \cite{schlechtweg2020semeval}.\\

Human languages constantly change due to cultural, technological, and social drift. Lexical semantic changes of human languages can materialize in the form of introducing/borrowing new words, or for the existing words can involve acquiring/losing some word senses. Computational methods for automated detection of semantic changes can be extremely helpful in the study of historical texts or corpora spanning a very long period of time, e.g., in the design of the OCR algorithm for text digitization, or in designing an information retrieval system incorporating the semantic changes \cite{tahmasebi2018survey}. Applications in the study of historical texts aside, the proposed methods detect lexical-semantic drift also in the same time period for different domains. This can be useful for 
compiling glossaries and specific training material in certain industries where new senses are introduced for words as compared to their standard usage e.g. to facilitate a more efficient training for new employees.\\

In the past decade, a variety of methods were introduced in the literature for automatic detection of lexical-semantic changes~\cite{tahmasebi2018survey}, where we only can refer to a subset of work, including but not limited to (i) co-occurrence-based methods~\cite{sagi2009semantic,basile2016diachronic}, (ii) embedding-based approaches~\cite{kimetal2014temporal,kulkarni2015statistically,asgari2016comparing,asgari2019life,asgari2020unisent}, (iii) topic-models-based ~\cite{frermann2016bayesian}, and (iv) alignment-based~\cite{bamman2011measuring} approaches. In this paper, we extend our recently introduced DomDrift embedding-based approach for the detection of semantic changes~\cite{asgari2020unisent} introduced for the extension of the computational analyses on 1000+ languages~\cite{asgari2017}. DomDrift works based on a comparison of relative distances of words in the embedding spaces of the source and the target domains. To increase the stability of detection in the DomDrift, we extend DomDrift to EmbLexChange by the following modifications: (i) instead of creating word profile against all common words between the source and target domain, we use only a subset of pivot words, which are frequent words with unchanged relative frequencies. (ii) We create multiple word profiles by resampling from a set of pivot words. We show that the EmbLexChange can reliably detect the lexical-semantic changes in English, German, Swedish, and Latin achieving an average accuracy of 0.686 as second best system of the competition where the first place system achieved an accuracy of 0.687.

\section{System overview}
\label{lab:overview}
\begin{figure*}[ht]
\centering
  \includegraphics[trim=0 0 0 0, width=1.01\textwidth,clip]{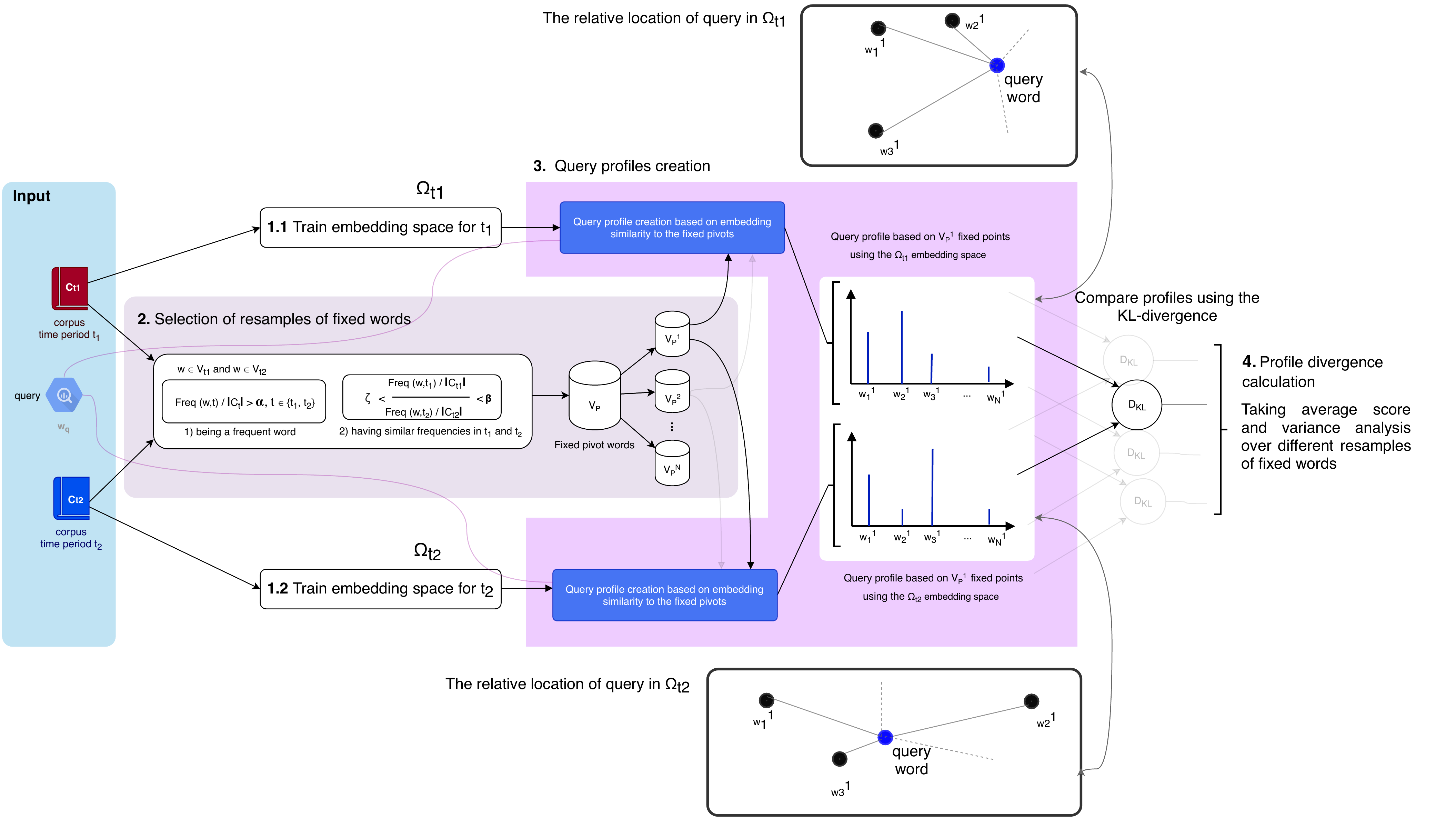}
  \caption{\label{fig:overview} The overview of EmbLexChange system for unsupervised detection of lexical-semantic changes. The steps are detailed in the \S\ref{lab:ebmlexchange}.}
\end{figure*}

Here we detail the steps of the EmbLexChange system, where the overview is depicted in Figure \ref{fig:overview}. The EmbLexChange framework
is developed based on the following assumptions: \\
\textbf{H1:} frequent words change at slower rates \cite{hamilton-etal-2016-diachronic}.
 \textbf{H2:} the relative frequency of unchanged words is not dramatically different in different time periods/domains. 
\textbf{H3:} changes of the word sense change the context and consequently alter the neighbors in the embedding space. Thus, the relative drift of a query word (a words which we target to investigate its lexical semantic change) with respect to unchanged words in the embedding space can characterize the lexical-semantic change. 

\subsection{EmbLexChange}
\label{lab:ebmlexchange}
\textbf{1. Training language-model-based embedding spaces:} The training of word embeddings using language modeling objective (e.g., skip-gram) has shown to preserve the syntactic and the semantic regularities in the vector space~\cite{NIPS2013_5021,pennington2014glove}. Semantic changes impact the neighborhoods in the embedding space ($H3$). Thus, the first step is to train embeddings separately for the text corpora in time period $t_1$ and $t_2$ (steps 1.1 and 1.2 in \ref{fig:overview}). In order to generate the embedding space $\Omega_{t}$, the only necessary resource is the raw text. For embedding creation, we use \texttt{fasttext}~\cite{bojanowski2017enriching} which leverages subword information within the skip-gram architecture. Using sub-word information minimizes the query terms Out-OF-Vocabulary ~\cite{bojanowski2017enriching}. The result of this step would be separate embedding spaces $\Omega_{t_1}$ and $\Omega_{t_2}$ for the time periods $t_1$ and $t_2$, $\Omega_{t_x}: V_{t_x} \rightarrow{}$ $\mathbb{R}^{h_{t_x}}, t_x \in \{t_1,t_2\}$ mapping their vocabulary $V_{l_{t_x}}$ to continuous vector representations in $\mathbb{R}^{h_{t_x}}$.\\ 

\noindent\textbf{2. Selection of fixed words and prepare pivot sets:} To measure the degree of semantic change for the given query words in $\Omega_{t_1}$ and $\Omega_{t_2}$, we need some fixed points, called pivot set $V_P$ comprising words with the property that their semantics are not dramatically changed and their relative positions in $\Omega_{t_1}$ and $\Omega_{t_2}$ remain almost constant (step 2 in \ref{fig:overview}). For this purpose, based on $H1$ and $H2$, we propose the use of frequent words with their relative frequency higher than $\alpha$ in both time periods of $t_1$ and $t_2$. Secondly, we filter this set by removing words whose relative frequency has changed between $t_1$ and $t_2$, resulting in $V_P$.  These fixed points are then used to create query profiles in $t_1$ and $t_2$. In order to increase the reliability and make variance analysis possible, we execute $N$ resamples each containing $M$ words from $V_P$: $V_P^{(1)}$,$V_P^{(2)}$,\dots,$V_P^{(N)}$.\\

\noindent\textbf{3. Query profiles creation:} In the next step, for each query word, we create $t_1$ and $t_2$ profiles based on the pivot resamples (step 3 in \ref{fig:overview}). The profile in time $t$ is an $l1$ normalized embedding similarity vector of the query to the terms in $V_P^{(i)}$:

\[
    P(w_q,V_P^{(i)},\Omega_{t})_j= \frac{\exp{(\phi cos(\overrightarrow{w_q}}, \overrightarrow{w_j^{(i)}}))}{\sum_{w_k^{(i)} \in V_P^{(i)}}  \exp{(\phi cos(\overrightarrow{w_q}}, \overrightarrow{w_k^{(i)}}))},  
  \]

where $V_P^{(i)}$ is the $i^{th}$ resample from the pivot set (created in step 2), $w_q$ is the query word, $w_k^{(i)}$ is the  $k^{th}$ word in the $i^{th}$ resample, $\overrightarrow{w}$ is the vector representation of word $w$ in the embedding space $\Omega_{t}$, $\phi$ is the temperature of the softmax function (used as a hyper parameter). Hence, for each $V_P^{(i)}$ we can create one profile in $t_1$ and one profile in $t_2$.\\

\noindent\textbf{4. Profile divergence calculation:} Next, for each resample $V_P^{(i)}$ we calculate the divergence between the profile in the time period $t_1$ and the time period $t_2$ using KL-divergence:

\[\lambda_{i}= D_{\mathrm{KL}}(P(w_q,V_P^{(i)},\Omega_{t_1})\|P(w_q,V_P^{(i)},\Omega_{t_2})) \]

We average the $\lambda_{i}'s$ over $N$ resamples as the measure of semantic change for the query word $w_q$. Since $ D_{\mathrm{KL}}$ does not have an upper-bound, we estimate an upper-bound based on $\lambda_{i}$'s on resamples of a large set of randomly selected words $V_{explore}$ including $V_P$ and a set of words with a change in their relative frequency. We draw $K$ resamples of size $M'$ words from $V_{explore}$ and calculate the \textbf{$\lambda_k$}
's ($\lambda$'s of words in $k^{th}$ resample of $V_{explore}$). We select the average of $90^{th}$ percentile over $K$ resamples as the upper bound and the average of $10^{th}$ percentile as the lower bound of the $\lambda$ to scale any calculated $\lambda_i$ for a query word to $\hat{\lambda_i}$ ($0\le\hat{\lambda_i}\le1$). Considering a threshold of $h$, we assign $\bar{\lambda} > h$ to the category of lexical semantic change, which can be adjusted as a hyperparameter on a validation set.

\section{Data}

The dataset used in this shared task includes corpora of English, German, Latin, and Swedish texts. For each language, the text corpora of two time periods are given. More details on the on the extact time frames and data sizes are provided in Table \ref{tab:data_summary}. 

\begin{table*}[ht]
\centering
\caption{\label{tab:data_summary} The English, German, Swedish, and Latin datasets used in the SemEval shared task.}
\resizebox{0.9\columnwidth}{!}{\begin{tabular}{l||c|c||c}
Languages & t1: time period ($\#$ tokens)       & t2: time period ($\#$ tokens)      & Reference  \\\hline\hline
English   & 1810-1860 (6.6M)& 1960-2010 (6.8M) & CCOHA \cite{alatrash2020CCOHA}      \\
German    & 1800-1899 (70.2M) & 1946-1990 (72.4M) & DTA, BZ, and ND corpora\\
 Latin & -200-0  (1.8M)  & 0-2000 (9.4M)   & LatinISE \cite{mcgillivray2013tools}   \\
Swedish     & 1790-1830 (71.1M)  & 1895-1903 (110.8M) & KubHist \cite{adesam2019exploring}  
\end{tabular}}
\end{table*}

\section{Experiment}
The goal of SemEval task 1 is to detect the words with a change in their semantics in the transition from the time period $t_1$ to the time period $t_2$ in English, German, Swedish, and Latin languages. We closely follow the steps described in $\S$\ref{lab:overview}. \\

\noindent\textbf{1. Language-model-based embedding setup:} We train \texttt{fasttext}~\cite{bojanowski2017enriching} embedding using the skip-gram architecture for each pair of language and time period separately. In the training of \texttt{fasttext}, we set the window size to $c=7$ and the embedding size of $d=100$. In the presence of a validation set, both $c$ and $d$ can be optimized as the hyper-parameters for each setting. It is known that a larger $c$ is favorable for semantics representation of words and a smaller $c$ for a syntax-related representation.\\
 
\noindent\textbf{2. Pivot resamples creation:} We firstly prepare a set of  frequent words existing in both $t_1$ and $t_2$ for each language considering the $\alpha$ (relative freq.) as a way to select the top $10\%-20\%$ frequent words. Next, we filter this set to keep the words with the property that the ratio of their normalized frequency is not substantially changed in $t_1$ and $t_2$, $\frac{2}{3}<\frac{freq_{t1}(w)}{freq_{t2}(w)}<\frac{3}{2}$ resulting in our $V_P$ set. Subsequently, we draw $N=10$ resamples from $V_P$ with the size of $M=5000$ for each language.\\

\noindent\textbf{3. Query profiles creation:} In the next step, as presented in $\S$\ref{lab:overview} step 3, for each query word we create $t_1$ and $t_2$ profiles for each $N=10$ pivot resamples as in the previous step.\\

\noindent\textbf{4. Profile divergence calculation:} Next, for each of $N$ resamples $V_P^{(i)}$, we calculate the $\lambda_i$
and scale them to $\hat{\lambda_i}$ using $K=5$ resamples of size $M'=5000$ words from $V_{explore}$. Subsequently, for the binary detection of changes, we apply different thresholds $h$ over the average of scaled divergences (the average of $\hat{\lambda_i}$'s $ = \bar{\lambda}$ ) of $V_P^{(i)}$'s and assign $\bar{\lambda} > h$ to the category of lexical semantic change.\\

\noindent\textbf{Evaluation:} For evaluation purpose, in the case of binary detection the accuracy metric is used to compare the given ground-truth and the predicted lexical semantic changes. For the ranking setting, we report both pearson correlation coefficient and Kendall-$\tau$ to measure the correspondence between calculated divergences ($\bar{\lambda}$'s) and the provided ground-truth scores.

\section{Results}

The results of EmbLexChange in the detection and the quantification of changes in the lexical semantics in English, German, Swedish, and Latin are provided in Table \ref{tab:result_summary}. After the competition, we had the chance to perform further optimizations of the hyperparameters leading to the current results, slightly improved from those submitted to the competition leaderboard. The EmbLexChange scores of the test set for all languages are available at http://language-lab.info/emblexchange/.\\

\noindent\textbf{Binary detection:} EmbLexChange could detect the semantic changes in English, German, Swedish, and Latin with the accuracy of $70.3\%$, $75.3\%$, $77.4\%$, $60\%$ respectively. The selected $h$ value, the thresholds to assign the positive or the negative class for each language, are also provided in Table \ref{tab:result_summary}.\\

\noindent\textbf{Ranking:} The Kendall-$\tau$ p-values for English, German, and Swedish show that there is a significant correspondence between the EmbLexChange scores and the ground truth scores in those languages. The Pearson correlation is also calculated for all languages, with an average of 0.306 over four languages. The case of Latin has been more challenging in both binary and graded prediction of lexical semantic change.\\

\begin{table}[ht]
\centering
\caption{\label{tab:result_summary} The summary of results for the detection and quantification of lexical-semantics changes in English, German, Swedish, and Latin in SemEval 2020.}
\resizebox{0.9\columnwidth}{!}{
\begin{tabular}{|l||c|c||c|cc}
\hline
\multicolumn{1}{|c|}{Languages} & Accuracy in binary detection & $h$ & Pearson correlation & \multicolumn{1}{c|}{Kendall-$\tau$} & \multicolumn{1}{c|}{Kendall-$\tau$-p-value} \\ \hline
English                         & 70.3                         & 0.4      & 0.314               & \multicolumn{1}{c|}{0.292}       & \multicolumn{1}{c|}{0.034}               \\ \hline
German                          & 75.0                         & 0.5      & 0.432               & \multicolumn{1}{c|}{0.315}       & \multicolumn{1}{c|}{0.009}               \\ \hline
Swedish                         & 77.4                         & 0.4      & 0.316               & \multicolumn{1}{c|}{0.383}       & \multicolumn{1}{c|}{0.0115}              \\ \hline
Latin                           & 60.0                         & 0.15     & 0.162               & \multicolumn{1}{c|}{0.161}       & \multicolumn{1}{c|}{0.222}               \\ \hline\hline
\multicolumn{1}{|c||}{Average}   & 70.7                       &          & 0.306               &                                  &                                          \\ \cline{1-2} \cline{4-4}
\end{tabular}}
\end{table}

\section{Discussion and Conclusions}
In this paper, we proposed EmbLexChange, a framework for the detection of lexical semantic changes in an unsupervised manner. We defined EmbLexChange as the divergence between the embedding-based profiles of word $w$ (calculated for a set of pivot words) in the source and the target domains (or between two time-frames). With the selection of pivot words by a resampling framework, we raise the reliability of this divergences. The underlying assumption of our method is that the changes in lexical semantics of word $w$ would affect its co-occurring words and subsequently alters the neighborhoods in the embedding spaces.\\ 

We showed that EmbLexChange can reliably detect lexical-semantic changes in English, German, Swedish, and Latin achieving the second place in the binary detection of semantic changes in the SemEval-2020. The detection of semantic changes in Latin has been more challenging than other languages. One reason behind this can be the imbalance of embedding training instances for Latin $t_1$
and Latin $t_2$ as well as the overall smaller corpora for Latin in comparison to the other languages (shown in Table \ref{tab:data_summary}). Another reason can be the split of time frames, where $t_2$ in Latin spans a large period of 2000 years.\\

The SemEval overall results show that EmbLexChange works better in the binary detection of semantic changes versus its performance in the ranking problem setting \cite{schlechtweg2020semeval}. However, we should note that the manual creation of ranking ground truth is a much more challenging task than the creation of binary classification ground truth. Thus, we believe that the classification results might be more reliable than ones for the ranking.\\

The EmbLexChange requires only the raw texts in the time-frames/domains of interest. Then the semantic changes can be detected based on the divergence between the embedding-based profiles of words between the source and the target domain. One advantage of using an embedding-based profile is that by increasing the window size in the embedding training we can move from syntactic changes toward semantic changes which can be investigated in more depth as a future direction.

\newpage
\bibliographystyle{coling}
{\smaller
}

\end{document}